\documentclass[preprint,12pt,numbers]{elsarticle}

\usepackage{amssymb}

\journal{Knowledge-Based Systems}

\begin{document}

\begin{frontmatter}



\title{A survey on bias in machine learning research}

\author[PG]{Agnieszka Mikołajczyk-Bareła}
\affiliation[PG]{organization={Gdańsk University of Technology},
            addressline={Gabriela Narutowicza 11/12},
            city={Gdańsk},
            country={Poland}}

\author[PG]{Michał Grochowski}


\begin{abstract}

Current research on bias in machine learning often focuses on fairness, while overlooking the roots or causes of bias. However, bias was originally defined as a "systematic error," often caused by humans at different stages of the research process. This article aims to bridge the gap between past literature on bias in research by providing taxonomy for potential sources of bias and errors in data and models. The paper focus on bias in machine learning pipelines. Survey analyses over forty potential sources of bias in the machine learning (ML) pipeline, providing clear examples for each.
By understanding the sources and consequences of bias in machine learning, better methods can be developed for its detecting and mitigating, leading to fairer, more transparent, and more accurate ML models.
\end{abstract}




\end{frontmatter}


\section{Introduction} \label{introduction}

Machine learning has become increasingly important in various fields, from healthcare, education, administration, finance to entertainment. However, as the use of machine learning grows, so does the risk of bias in the data and models used. First of all, bias in machine learning is one of the main contributors to incorrect operation of systems, due to obvious errors in reasoning. Bias in machine learning can have serious consequences, such as perpetuating societal inequalities, discriminating against certain groups. For example, in criminal justice, biased data can lead to unfair decisions, such as disproportionately targeting certain demographics or falsely identifying someone as a criminal \cite{juliaangwin_2016}. In healthcare, biased data can lead to inadequate diagnoses, mistreatment, or poor patient outcomes, as was the case with the skin cancer detection algorithm that showed significant bias towards lighter-skinned patients \cite{esteva2019guide}. As such, it is crucial to identify and address bias in machine learning to ensure fair and equitable outcomes.

This survey provides an overview of the current state of knowledge regarding bias in machine learning research, including the review of biases at different stages of research and contemporary approaches to bias detection and mitigation. 
Bias can be broadly defined as "a systematic deviation of results or inferences from the truth or processes leading to such deviation" (Choi et al. \cite{k2014bias}). In machine learning, bias is often referred to as "a \textbf{systematic error} from erroneous assumptions in the learning algorithm" (Mehrabi et al. \cite{mehrabi2021survey}). Bias can be found in all areas of research. It can interfere with research project at any stage, including the beginning (e.g., literature review or data collection), the middle (e.g., the model training), and the end (e.g., evaluation and closure of a research project) (Choi et al. \cite{k2014bias}). Avoiding bias demands ongoing attention and awareness from all project members. However, it is natural that even with such efforts, errors can still occur. Nevertheless, to ensure that bias does not interfere with our research, we must first be aware of its existence.  


To ensure that machine learning models are not biased, researchers have developed various approaches to detect and mitigate bias in data and models. Some of these approaches include fairness metrics, debiasing techniques, and explainability methods. Fairness metrics provide quantitative measures of fairness in models by evaluating their performance across, for example, different demographic groups. Debiasing techniques aim to reduce the impact of bias in the data by removing or balancing biased features or samples. Explainability methods aim to provide insights into how the model makes its decisions, allowing for the identification and correction of any biases.

To detect and mitigate bias, we first need to clarify definitions and establish the scope of the topic.
This article aims to provide an overview of the current state of knowledge regarding bias in machine learning, including the review of biases at different stages of research and contemporary approaches to bias detection and mitigation. By understanding the sources (or causes) and consequences of bias in machine learning, we can develop better methods for detecting and mitigating bias, leading to fairer, more transparent, and more accurate machine learning models.

In the "Related Works" section, we present how our research compares to other surveys that focus on bias in both the well-defined and established field of bias in research and the developing field describing bias in machine learning.  In the next section, "Sources of bias at different stages of research," we define, categorize, and present potential sources of bias in the machine learning pipeline. To help readers better understand each bias, we provide a clear example for each. Finally, we offer a brief overview of methods for discovering and mitigating bias.

\section{Related works}

In recent years, there has been an increasing interest in addressing the growing problem of bias in machine learning. Researchers and industry leaders alike have recognized the serious consequences of perpetuating societal inequalities, discriminating against certain groups, or producing inaccurate results. As a result, several surveys on fairness and bias in machine learning have emerged, and we have taken into account the most impactfull works while preparing this survey. 

However, machine learning research commonly focuses on fairness when addressing bias, often neglecting the initial description of bias as a "systematic error." We believe that this difference has resulted in a separation between prior and current research on bias, with an insufficient understanding of its origin and harmful consequences. The absence of identification of genuine causes of bias poses difficulties in attempting to minimize its effects. Our aim in this paper is to address this separation by investigating past and current research on bias in data and models.

\subsection{Bias in research}

To begin with, we analyzed prior literature on bias in research (without specifically focusing on machine learning). Choi et al. \cite{k2014bias} catalogued and described 109 biases that can be found in a research study. They defined bias as "systematic errors that decrease the validity of estimates, and does 
not refer to random errors that decrease the precision of estimates". Therefore, unlike random error, bias cannot be eliminated or reduced by an increase in sample size. 
In previous widely accepted works, bias was described as a result of flaws in the following stages of research \cite{k2014bias, sackett1979bias}: (1) Literature review (errors in reading-up on the field), (2) Study design (errors occurring as a result of faulty design of a study), (3) Study execution (errors in executing the experimental maneuver), (4) Data collection (a flaw in measuring exposure or outcome that results in differential quality or accuracy of information between populations), (5) Analysis (errors in analyzing the data), (6) Interpretation of results. (error that arises from inference and speculation), and (7) Publication (an editorial predilection for publishing particular findings). 

Maclure and Schneeweiss \cite{maclure2001causation} focused on the bias in the epidemiology, although the definitions are applicable to other fields as well. They showed that \textit{bias is caused} by using the theory of causal diagrams, although concluding that sources of bias often cannot be treated as confounders. Similarly to the stages presented by Choi et al. they presented different steps, presented as layers of \textit{"lenses"} and \textit{"filters"} in  \textit{The Episcope} device (epidemiologist’s “telescope”
for observing populations). Each layer of filters/lenses in The Episcope represents another research stage (distinct domain) where certain types of biases operate, potentially adding additional distortions to the association of interest \cite{maclure2001causation}. 

Smith and Noble \cite{smith2014bias} outlined types of bias across different research designs, and proposed strategies to mitigate the bias. They defined five types of research bias which can be introduced on different phases of research: design bias (occurs when the study design was poor and there was a mismatch between objectives and methods), selection/participant bias (relates to both the process of recruiting participants and study inclusion criteria), data collection bias and measurement bias (when a researcher's personal beliefs influence the way information or data is collected), analysis bias (when researcher naturally look for data that confirm their hypotheses and personal experience, thus rejecting data not corresponding to them) and publication bias (In quantitative research, for example, studies are published significantly more if they report statistically significant results). \cite{smith2014bias}.

Delgado-Rodríguez and Llorca, also showed that biases can be classified by the research stage in which they occurred \cite{Delgado635}. However, they focused only on the selection, information, and confounding bias.

Using above-mentioned works that are well-established in the field, we proposed research stages that are adequate for machine learning research. Proposed stages we adapted to the typical ML pipeline by focusing more on data and adding model selection and training stage. Final stages are described in the Section \ref{sec.sources}. 

\subsection{Sources of Bias in Machine Learning}
Recent research on bias in machine learning has primarily focused on fairness, rather than the sources or causes of bias, with a few exceptions. Mehrabi et al. \cite{mehrabi2021survey} conducted a survey on bias and fairness in machine learning and identified two main sources of unfairness: biases arising from the data and algorithms. These biases can affect users, which in turn can impact data and other algorithms, resulting in a self-reinforcing feedback loop. This feedback loop consists of three main groups of biases. The first group, called \textit{Data to Algorithm}, describes situations where biased data is used to train ML models, leading to biased outcomes. The second group, \textit{Algorithm to User}, refers to a scenario where the user is biased due to algorithmic outcomes. The last group, \textit{User to Data}, is related to the way in which users gather or process data. In this review, we extend Mehrabi's categorization by aggregating more recent works and focusing on systematic errors. We provide a more granular review of biases, presenting not only the potential sources of bias but also situating them in the ML research pipeline. While Mehrabi's focus was limited to the data and models (biases arising from the \textit{data collection}, \textit{data analysis}, and \textit{model selection} stages), we also show that biases can occur between users (\textit{User to User}), and that data can bias the user (\textit{Data to User}).

For instance, biases can arise at different stages of research. During the literature review stage, the biases of the researchers or authors can inadvertently influence their work, resulting in \textit{User to User} bias. Biases can also stem from the data collection stage, such as \textit{User to Data} bias (when annotating data) and \textit{Data to Algorithm} bias (when biased data is used to train the model). In the data analysis stage, the most common bias is \textit{Data to User} bias (e.g., mistaking correlation in data with causation). Similarly, in the model selection and training stage, the most common bias is \textit{Algorithm to User} bias. Finally, during the interpretation and publication of results, we observe the bias of \textit{Algorithm to User} (when a biased algorithm produces biased results for the user) and \textit{User to User} (when a biased user presents biased results to others).

We provide a summary of the comparison between Mehrabi et al.'s categorization and our bias categories in Table \ref{tab:merhabi}.

\begin{table}[ht]
\caption{Comparison between Merhabi et al. and our bias categories}
\label{tab:merhabi}
\centering
\begin{tabular}{@{}ll@{}}
\textbf{Stage} & \textbf{Most common bias} \\
Literature review & User to user \\
Data collection & User to Data, Data to Algorithm \\
Data analysis & Data to User \\
Model selection and training & Algorithm to User \\
Interpretation of the results & Algorithm to User \\
Publication & User to User \\
\end{tabular}
\end{table}

Cirillo et al. \cite{cirillo2020sex} presented a review of different biases in machine learning. The first and primary difference between our works is the definition of the word "bias." In our paper, the word "bias" is defned as a systematic error that might come from different sources. Although the definition of bias is not directly mentioned in Cirillo et al.'s work, we can infer from the content that they refer to an inclination or prejudice towards/against one person or group, especially in a way that is considered unfair. Cirillo et al. defines desirable and undesirable biases that can be found in data and models. By adopting a separate definition of bias as a "systematic error," we cannot distinguish categories related to desirabile bias since every systematic error is undesirable and should be eliminated. Not every difference that occurs between genders, ethnic groups, or other populations is incorrect (as in the definition of bias), even when it is unfair (as in the definition of fairness). The authors define desirable bias as bias that takes advantage of gender differences to select the best possible treatment process, therapy, or diagnosis ("\textit{A desirable bias implies taking into account sex and gender differences to make a precise diagnosis and recommend a tailored and more effective treatment for each individual.}"). A similar definition was presented in the paper by Pot et al. \cite{pot2021not} who propose to understand bias as a social problem and analyze its causes and implications through a framework of equity in healthcare. They suggest that not all biases are "bad" and propose to think of some biases as "good" and desirable, as they can help to overcome existing inequities in healthcare. 
In our review, \textit{fully justified differences between populations are not called bias}, as they are merely characteristics of a population. If the differences are real, well-studied, on an appropriate population, without making any errors -- in other words, if such differences exist, \textit{they are not systematic errors}, no matter how prejudicial or unfair they are to the studied population. Such "differences" (one could also say, features) simply reflect the nature of the problem being analysed.

Referring to the bias categories outlined by Cirillo et al., we would like to emphasize that all the biases mentioned in our paper are undesirable biases because they are defined as \textbf{errors that we aim to avoid}. Eliminating or mitigating systematic biases will always result in improved system performance, benefiting the user (e.g., the patient).

Regarding the systematic nature of biases, Cirillo et al. lists six main types, which they refer to as sources: Historical bias, Representation bias, Measurement bias, Aggregation bias, Evaluation bias, and Algorithmic bias. This method of categorizing biases highlights that the authors view inequalities between populations as being rooted in biases. They list biases as sources of bias. It is likely that they mean biases (here as prejudices) have their roots in the listed biases (here as systematic errors).

We have divided a typical research-based machine learning project into stages. In each of these, different types of biases can occur. Our stages are consistent with the sources outlined by Cirillo et al. These sources of bias fit into the three stages of an ML project: data collection, model preparation, and analysis of results. Below, we have provided a condensed summary of the proposed ML project stage descriptions, with subsections referencing Cirillo's work.

\begin{itemize}
    \item \textit{Data collection }
    \begin{itemize}
        \item \textit{Measurement bias} -- occurs when measured data are often proxies for some ideal features and labels.
        \item \textit{Representation bias} -- occurs when certain parts of the input space are underrepresented.
        \item \textit{Historical bias} -- arises even if the data is perfectly measured and sampled, when the world as it is leads a model to produce outcomes that are not desired.
    \end{itemize}
    \item \textit{Model selection and training.}
    \begin{itemize}
        \item \textit{Algorithmic Bias} -- 
occurs when bias is introduced in the algorithm consciously or unconsciously in ad-hoc solutions.

        \item \textit{Aggregation bias} -- arises when a one-size-fits-all model is used for groups with different conditional distributions.
        \end{itemize}
    \item \textit{Interpretation of the results.}
    \begin{itemize}
        \item \textit{Evaluation bias} -- occurs when the evaluation and/or benchmark data for an algorithm does not represent the target population.
        \end{itemize}
\end{itemize}




Hove and Prabhumoye's \cite{hovy2021five} paper discusses the sources of bias in natural language processing applications and possible counter-measures, mainly focusing on demographic factors. Their definition of bias refers to the discrepancy between the intended and actual distributions of labels and user attributes during system training and application \cite{shah2020predictive}. The paper identifies three main sources of bias related to the data used to train models: data selection, annotation, and input representations. The authors also discuss how bias can originate from the models themselves and the overall research design. In our survey, we similarly explore the sources of bias in machine learning, but with some variations. We recognize literature review and publication bias as common stages in any type of research, and we divide the data collection and analysis into two stages instead of three. Additionally, we include the model selection and training stage, which encompasses inference and is partially overlooked by Hove and Prabhumoye's review. While Hove and Prabhumoye provide a concise representation of potential biases in NLP, they omit several recently reported biases, as well as well-established causality biases. Our survey aims to extend this research and present a comprehensive review of the sources of bias in machine learning.

\section{Sources of bias at different stages of research} \label{sec.sources}


\begin{figure}[!htb] 
\centering 
  \includegraphics[width=1.0\textwidth]{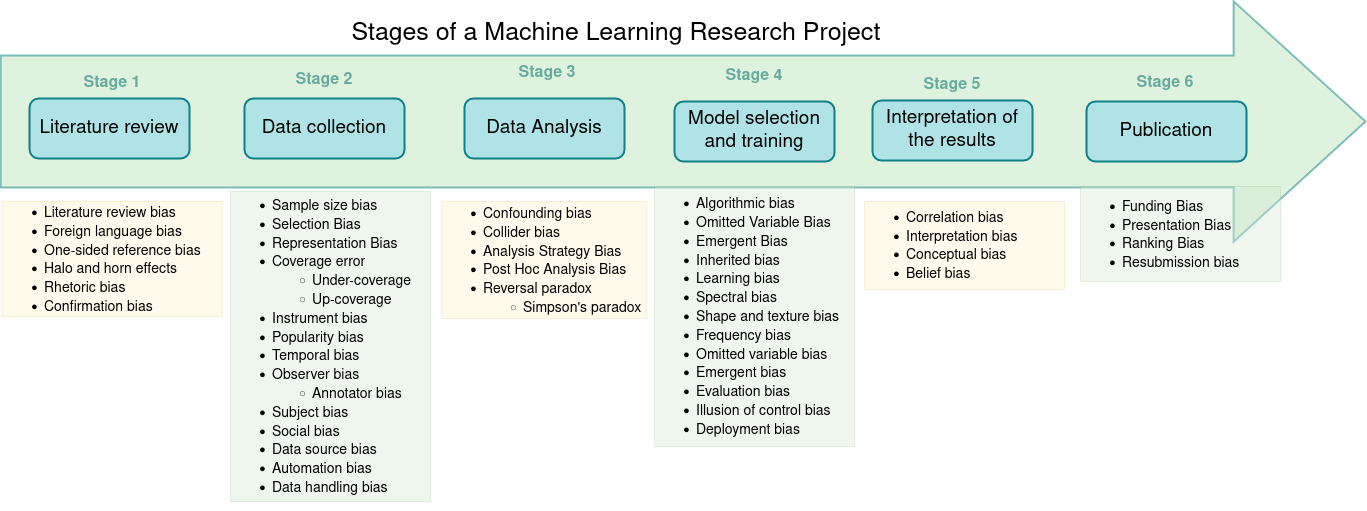}%
  \caption{Stages of a machine learning project and potential sources of bias} \label{Fig.stagesofbias}
\end{figure}

In this section, we have presented a concise overview of over forty potential sources of bias that can occur at different stages of the typical machine learning pipeline. To aid in understanding these stages, we have included a visual representation in Figure~\ref{Fig.stagesofbias}.

We have identified six distinct stages in a typical machine learning research project:
\begin{itemize}
\item \textit{Literature review.} A literature review involves a thorough analysis and examination of previously published works related to a specific topic. In the context of machine learning, this includes scholarly papers, open-source repositories, existing projects, and programming libraries.
\item \textit{Data collection.} This stage involves gathering or measuring targeted variables in a predefined system to answer relevant questions and evaluate outcomes.
\item \textit{Data analysis.} Data analysis and exploration are essential steps of any research project. This stage involves inspecting, cleaning, preparing, feature selection, and exploring data to discover meaningful information about the investigated process or relation. Exploratory data analysis is often employed to summarize the primary characteristics of data collections, often using statistical graphics and other data visualization methods. The results of data analysis shape the future directions of the subsequent stage: model selection and training.
\item \textit{Model selection and training.} This step involves choosing the most appropriate model for the existing data and problem, designing the training pipeline, selecting the input features and hyper-parameters, finding the model's parameters by training, and validating it to achieve planned efficiency. This process heavily relies on the knowledge gathered from the previous data analysis stage. Additionally, we include biases in inference as a part of this stage.
\item \textit{Interpretation of results.} This stage covers analyzing the results of experiments, including comparison with state-of-the-art or ground-truth, interpreting the results, and forming conclusions about the experiments.
\item \textit{Publication.} The publication stage essentially focuses on sharing the results of experiments, code, methodology, and project summary.
\end{itemize}

\subsection{Literature Review}
The methodology of reviewing the state-of-the-art can also introduce biases that impact the final study results, although to a lesser extent than other stages in the model design process. \textit{Literature review bias}, also known as \textit{reading-up bias}, refers to errors in comprehending the field \cite{aleu2020assessing,k2014bias}. One commonly known type of literature review bias is \textit{literature search bias}, which results from an incomplete search due to poor keyword selection or search strategies or a failure to include unpublished reports and hard-to-reach journals \cite{k2014bias}. Ignoring this bias may lead to repeating failed experiments or addressing problems that have already been well-defined and researched in the past. In the machine learning community, \textit{literature review bias} can be introduced by using a different dataset splits or training data shuffle in method's comparison, or reporting false results by testing model performance on the validation set instead of the separate test set or testing too little data.

Sometimes, the literature search can be biased towards a single language, leading to \textit{foreign language exclusion bias} \cite{k2014bias}, where publications in foreign languages are ignored. This exclusion may result in a significant bias in selection \cite{dubben2005systematic}. The web's vastness also poses challenges in finding relevant articles, which may result in both the \textit{halo and horn effects}. Poorly written or badly structured papers can leave an impression of low-quality research, decreasing trust in achieving results (horn effect). Even with outstanding achievements, such reports might remain unnoticed, while some average articles might be over-glorified due to the great impression of the journal or authorship (halo effect).

Another type of literature review bias is \textit{one-sided reference bias}, which occurs when researchers restrict their references to only those studies that support their position \cite{gotzsche1987reference}. Researchers may unintentionally introduce \textit{rhetoric bias} when they try to convince the reader without scientific facts or reason \cite{k2014bias}. These two types of bias are closely related to what is often discussed in psychology as \textit{confirmation bias}, defined as \textit{the tendency to interpret new evidence as confirmation of one's existing beliefs or theories}\footnote{definition from Google's English dictionary provided by Oxford Language}. This bias means that the literature review or any other information searched for, interpreted, or analyzed, is systematically favored by the biased researcher towards their position or hypothesis \cite{oswald2012confirmation}. Confirmation bias is not strictly related to machine learning and can be found at any research stage.

\subsection{Data Collection}

The quality of the dataset used for synthesis of any model, especially data-driven model has a significant impact on the final results. The phrase \textit{garbage in, garbage out} was first used in 1957 in the article \textit{Work with new electronic 'brains' opens field for army math experts} \cite{mellin1957work} to refer to the software development process. Since then, it has become a famous saying within the machine learning (ML) community, explaining that even if the model is well-designed and correctly implemented, unsatisfactory results may occur if the data used is of low quantity, low quality, incorrect, biased, or noisy.

In real-world projects, data preparation is usually a tedious, lengthy and costly process. During this procedure, unwanted bias may accidentally be inserted into the dataset. This section highlights the three substages involved in the dataset preparation stage: design of data acquisition, execution, and collection.

One of the key factors that influence data quality is the general design of the data acquisition process. To minimize the potential impact of biases, it is essential to follow a standardized protocol for acquiring and gathering data. In many cases, biases are introduced during the initial step of data collection, i.e., the design of the data collection process. However, the absence of a standardized guideline for collecting data in machine learning has made it difficult to ensure uniformity in data collection. Although some commonly used datasets follow standards required for top machine learning conferences and journals, they may still be biased due to the way the data was collected. 

One of the most common problems is a sample size referred to later as dataset size. It is well known that deep learning algorithms require vast amounts of data to be effectively trained. A small dataset makes training more challenging and makes the proper data randomization process harder. This problem is called a \textit{sample size bias}.  

The flawed design of datasets is not uncommon in the field of machine learning. It can frequently be observed in crowdsourced datasets \cite{zheng2017truth} and even in widely accepted medical datasets \cite{gregory2012research}. Data may be inadvertently collected in a manner that introduces \textit{selection bias}. Selection bias is defined as a deviation of data from the truth resulting from how samples were collected. It may arise when a) the sampling frame is incomplete or inaccurate, b) the sampling process was nonrandom, or c) some targets were excluded from data collection.

\textit{Sampling Bias} (also known as Representation Bias) is a bias where data is collected in a manner that not all samples have the same probability of being selected in the study, resulting in unequal representation \cite{mehrabi2021survey}. For instance, the Open Litter Map \cite{lynch2018openlittermap}, which is currently the most comprehensive dataset of images containing litter, suffers from a significant \textit{sampling bias}, which is also known as \textit{sampling frame bias} and \textit{representation bias} \cite{mehrabi2021survey}.  The main objective of the project was to gather as much data as possible, instead of focusing on its quality. Waste images were collected and annotated by anonymous users who wanted to help the environment. However, since the website and application were initially only available in English, most users were from the UK, with some from the USA. If the problem of sampling bias in the dataset went unnoticed, it could lead to false assumptions about the study's phenomenon. For example, one might assume that "there is more litter in the UK than in India," whereas the real reason could be that people in the UK uploaded more images than people in India. Lack of geographical diversity is a common issue in many datasets used worldwide \cite{mehrabi2021survey}. Another example of sampling bias is the prediction of who is more likely to commit a crime based solely on data of those who were arrested \cite{chouldechova2018frontiers}. This approach is flawed, as it is impossible to gather data on all crimes and only data on arrests is available. As a result, the analysis only covers offenders who have been caught and not those who have not, and yet committed a crime.

Another problem of bias in data is a \textit{coverage error}, which occurs when the sampling frame is flawed, leading to a mismatch between the target population and the sample population \cite{wolter1986some}. There are two types of coverage error: \textit{under-coverage} (Non-coverage bias and \textit{up-coverage}). Non-coverage bias, arises when certain data samples are impossible or challenging to obtain \cite{k2014bias}. On the other hand, up-coverage bias occurs when the same data sample is erroneously considered as two distinct samples. For instance, in a study examining the spread of a virus, if a single patient's data is recorded twice, it would result in an up-coverage error. Conversely, if some patients' data is not recorded, it would be an example of under-coverage error. Coverage bias is difficult to define in machine learning, but it appears to be more prevalent in surveys or tabular data, such as response bias or illegal immigrant bias. The impact of coverage bias can be significant, and it can lead to skewed results, inaccurate inferences, poor generalisation and false conclusions.

\textit{Nonrandom bias} is a type of bias that occurs when the selection process is affected by the human choice, e.g., when sampling is nonrandom \cite{mclachlan1984method,k2014bias}. A classic example of nonrandom bias was described in the book \textit{How to lie with statistics} by Huff Darrell \cite{huff1993lie}. It shows an example of sending a questionnaire about loving surveys and gathering the answers of those who responded. This method of survey collection created a nonrandom bias, as people who enjoy responding to surveys are more likely to complete them than those who dislike them. To avoid nonrandom bias, researchers should select a representative sample of the population and ensure that all participants complete the survey.

\begin{figure*}[!htb]
\centering
\footnotesize

  \includegraphics[width=0.25\textwidth]{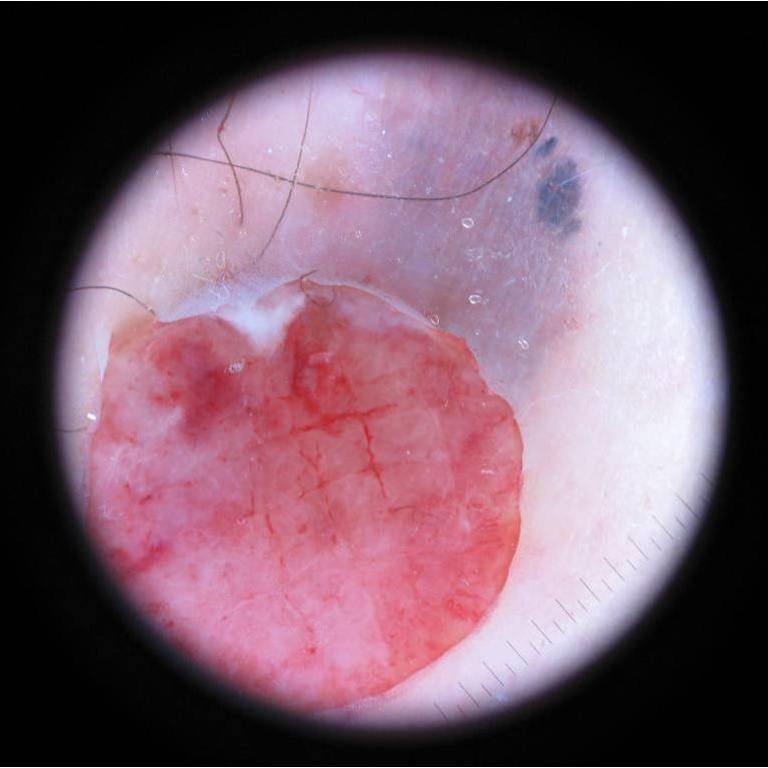}
  \includegraphics[width=0.25\textwidth]{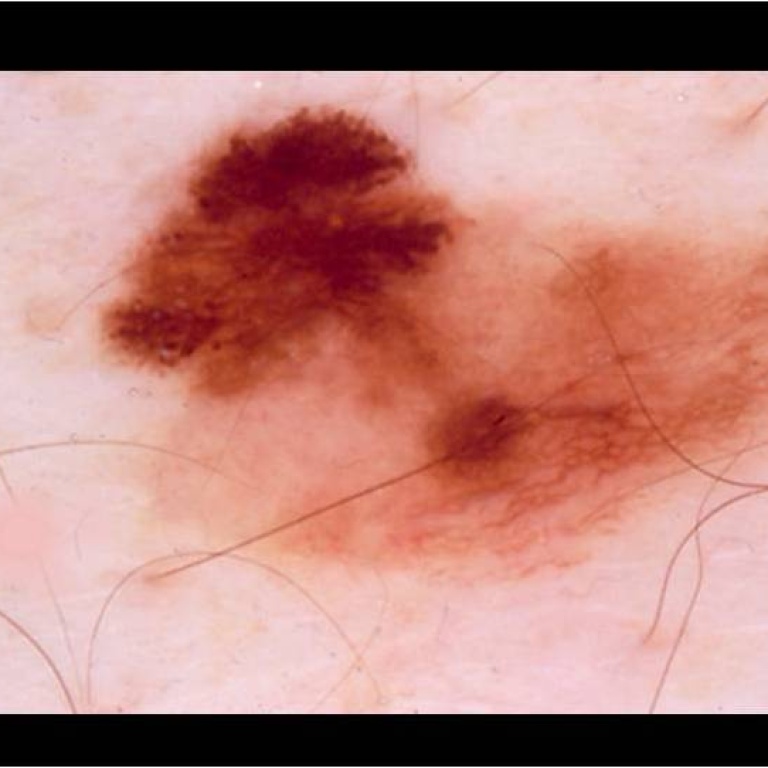}
  \includegraphics[width=0.25\textwidth]{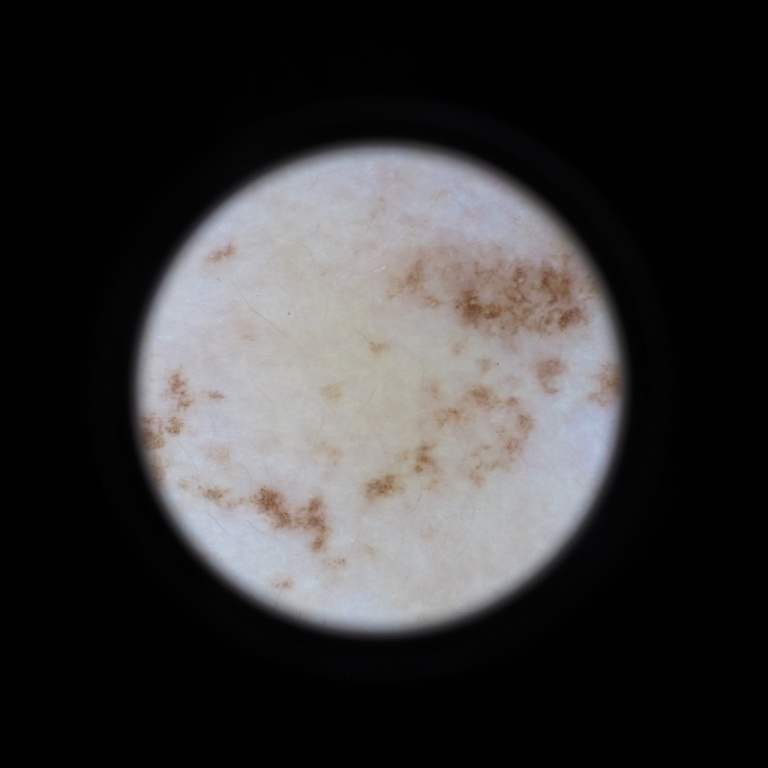}
  
 A: A dermoscopic image of skin lesions with black frame \\

\includegraphics[width=0.75\textwidth]{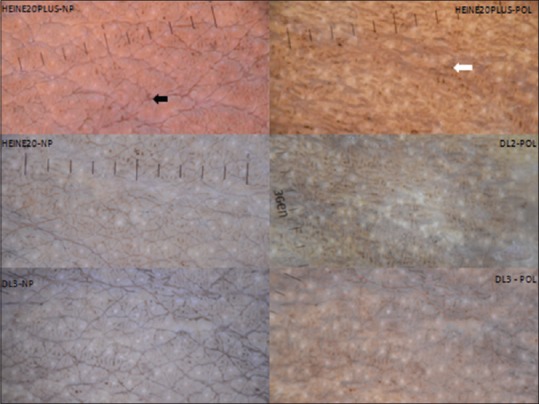}

B: Dermatoscopy of Acanthosis nigricans over the neck. Differences between common dermatoscopes (source: \cite{nirmal2017dermatoscopy}) \\

  \caption{Examples of instrument bias in dermoscopy presenting dermoscopic images.}
  \label{fig.bias.frames}
\end{figure*}

\textit{Instrument bias} results from imperfections in the instrument or method (including habits, experience) used to collect or manage the data \cite{he2012bias}. Devices used to collect data can strongly affect the learning algorithm. For instance, Nirmal et al. analyzed the typical differences in dermatoscopes - medical apertures used to observe and take images of skin lesions \cite{nirmal2017dermatoscopy}. They found that some dermatoscopes can better show certain structures, while others allow for better visualization of different features \cite{nirmal2017dermatoscopy}, as shown in Figure~\ref{fig.bias.frames}. Using a poor-quality dermatoscope that misses visible structures of a skin lesion with malignant characteristics, or alters the white balance, would result in instrument bias. Such a device could hinder the classification of the lesion into malignant or benign by missing valuable information. 

Additionally, a particular dermatoscope used to collect skin lesion images could modify the actual skin lesion by adding a black frame. Such a case was observed by Bissoto et al. \cite{bissoto2019constructing}, as the models trained on images without skin lesion achieved similar accuracy in skin lesion classification as original images. The subject was more deeply investigated by Mikołajczyk et al., who showed that black frames are not only correlated with malignant class, but also with other dermoscopic artifacts like ruler marks as well. These examples of instrument bias are presented in Figure~\ref{fig.bias.frames}.

In some cases, even the time of data collection is essential. The two examples are \textit{popularity bias} and \textit{temporal bias}. Temporal bias is defined as "systematic distortions across user populations or behaviors over time. "\cite{olteanu2019social}. In other words, temporal bias might undermine the validity of predictions by overemphasizing features close to the outcome of interest \cite{yuan2021temporal}. Yuan et al. \cite{yuan2021temporal} presented example of temporal bias in medical diagnosis during the discovery of lyme disease in 1976. Lyme disease shows following symptoms in that order: (1) an initial bite, (2) an expanding ring rash, and (3) arthritic symptoms \cite{steere2016lyme}. Researchers only studied patients who had already developed arthritis (stage 3), overlooking the significance of a tick bite (stage 1) and the ring rash (stage 2) that precedes it. A doctor examining a patient with a tick bite would miss the possibility of disease until further symptoms developed, and a predictive model focused on ring rashes would report false negatives for patients who had yet to develop the rash. These errors result from an incomplete understanding of the full range of symptoms \cite{yuan2021temporal}.

Meanwhile, popularity bias stems from increased public interest in a research subject \cite{ciampaglia2018algorithmic}. Recommendation systems \cite{abdollahpouri2019unfairness} have a common problem: popular topics (items, movies, books) are recommended more often, whereas less popular are recommended less frequently or never \cite{abdollahpouri2019managing}. As a result, the popular items became even more popular, and niche items got lost in the sea of propositions. Both popularity bias and temporal
bias represent a mismatch between the data used in the study
and the data used during the inference.

Next, there is an \textit{observer bias}, sometimes called a \textit{research bias} or an \textit{experimenter bias}. It owes the name to its definition: it tends to observe what the observer wants to see \cite{mahtani2018catalogue}. A famous example of observer bias is Cyril Burt's research on the heritability of IQ. Burt, an English educational psychologist, believed that children with a higher socioeconomic status were more intelligent on average. His research led to the creation of a two-tier educational system in 1960s England, which sent middle- and upper-class children to elite schools, while working-class children were sent to less desirable schools \cite{burt1943ability}. Currently, he is well-known as a researcher who falsified his work \cite{lamb1992biased,fletcher1991science, jensen1980bias}. 

In data-driven systems, observer bias might appear when annotators use personal, subjective opinions to label data, resulting in incorrect annotations. Depending on the annotated data, it might be tough to differentiate emotional thoughts from objective observations. An example might be a sentiment analysis, where annotators must decide if the sentence (written or spoken) has a negative, neutral or positive meaning \cite{kiritchenko2018examining}. In some cases, the annotation process is even more advanced: e.g., in emotion recognition, annotators have to divide spoken conversation into seven different emotions: \textit{neutral, sad, angry, happy, surprised, fear, and disgust} \cite{koolagudi2012emotion}. Even in cases where annotators do not have to tag emotions, they might still let their habits into the procedure.
An example might be adding punctuation to the spoken text, which is particularly useful in punctuation restoration tasks \cite{yi2020focal, moro2017prosody}. The exact page of text is often tagged differently by annotators \cite{bohavc2017text} even when following the guidelines. Some people have preferences to stay with longer, complex sentences, whereas others prefer to keep them short \cite{bohavc2017text}. When an annotator uses his prejudice to label that, this sub-type of subjective bias is called an \textit{annotator bias} \cite{hellstrom2020bias}.
In many cases, developing a well-designed data acquisition pipeline and annotation guidelines can help eliminate the above-mentioned biases. 

\begin{figure*}[!htb]
\centering
\includegraphics[width=\linewidth]{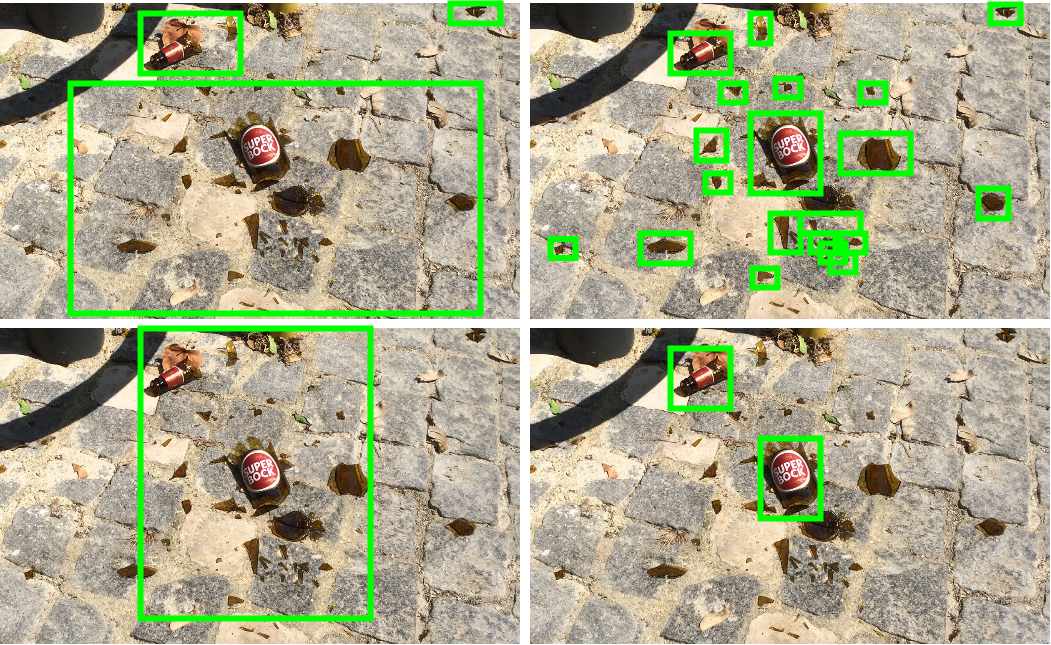}
  \caption{Labeling consistency problem. Waste detection example: how to label broken glass bottles? Annotated image come from the TACO dataset \cite{proencca2020taco}}
  \label{fig.data.labeling_consistency}
\end{figure*}

The subjectivity bias might also result in the inconsistencies in the data labeling process, which usually lead to bad model's performance during the training. For instance, leaving the decision of how to label broken glass in a waste detection problem to annotators, as shown in Figure \ref{fig.data.labeling_consistency}, can result in each annotator labeling the image differently, leading to inconsistently labeled dataset.

This problem is usually handled with additional coefficients to measure the agreement between different annotators, i.e., Cohen's Kappa score \cite{artstein2017inter} or by having two or more annotators label the data independently. This can help identify any discrepancies and ensure that each annotator labels the data consistently. 

\begin{figure*}[!htb]
\centering
\includegraphics[width=\textwidth]{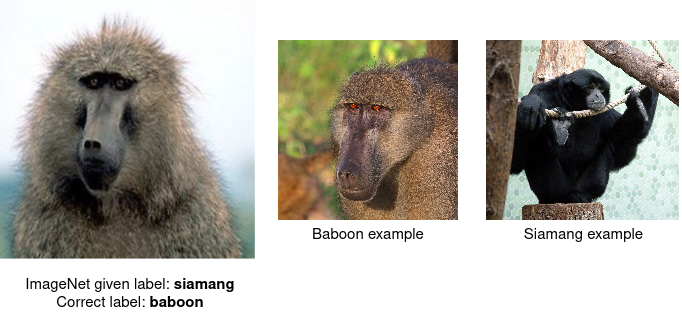}
\caption{Mislabeled samples problem. ImageNet example discovered by \cite{pleiss2020identifying}.}
\label{fig.data.mislabeling}
\end{figure*}

Usually, preparing the annotation guidelines is a continuous development process, not a one-time task. Inconsistencies can be discovered during the annotation process,  or even after, when training results are unsatisfactory (e.g. due to human error as presented in fig. \ref{fig.data.mislabeling}). Unfortunately, reaching out to annotators is not always feasible, especially when using open-source public benchmarks. 

Research has shown that even the most commonly used datasets are often mislabeled. 
Pleiss et al. reported an average labeling error rate of 3.3\% \cite{pleiss2020identifying}, while Li et al. found that the error rate reached over 17\% for the WebVision50 data set, which contains two million images scraped from Flickr and Google Image Search \cite{li2017webvision} (a website has been created to display all discovered mislabeled samples \footnote{Label errors in commonly used ML benchmarks: https://labelerrors.com/}). The problem is apparent also in other domains. The most popular multi-domain task-oriented dialog dataset MultiWOZ was proven to have severely inconsistent annotations, i.e. depending on the domain, total number of samples that needed correction varied from 2.1\% to 86.2\% \cite{qian2021annotation}. In addition, they spotted an entity bias, e.g., “cambridge” appeared in 50\% of the destination cities in the train domain\cite{qian2021annotation}. Removing mislabeled data, even at the cost of losing some data points, has been shown to improve the model's error rate (5-10\%). Furthermore, mislabeled samples can be identified automatically by measuring their contribution to generalization with margins. 
To address this issue, some approaches have been proposed, such as automated validation of label consistency in named entity recognition datasets. For instance, Zeng et al. \cite{zeng2021validating} proposed a method for validating label consistency automatically.

However, the observer (e.g., annotator) is not the only one who can add bias to the data. Another type of bias is \textit{observee bias}, widely known as \textit{subject bias}. It refers to the inaccurate data provided by the subjects. Choi et al. recognized multiple types of such bias, including subjects' preferences that give wrong information intentionally or unintentionally. It might alter the data whenever the subject is the primary provider of the data (interviews, questionnaires, reports, etc.) \cite{k2014bias}. 

Collected data might also be biased by some inequities pre-existing in our society, like stereotypes or historically disadvantaged groups. Those are sometimes called \textit{cultural biases}, \textit{social biases} or \textit{stereotypical bias}, and are mostly found in various text corpora \cite{kaneko2022gender}. Nangia et al. define nine types of stereotypical bias: race, gender (gender identity or expression), socioeconomic status (occupation),  nationality, religion, age, sexual orientation, physical appearance, and disability bias \cite{nangia-etal-2020-crows}. According to Nangia et al., a sentence is stereotypical when an advantaged group (e.g., a high socioeconomic status) is associated with a pleasant attribute (e.g., \textit{People who live in a mansion are smart.}) or a disadvantaged group with an unpleasant adjective (e.g., \textit{People who live in trailer parks are careless.}).

Next, a bias connected to the data acquisition process includes \textit{data source bias} \cite{k2014bias}, including competing death bias, family history bias, and spatial bias \cite{k2014bias}. Those biases are primarily reported in medical papers. Since ML is often used to support work in hospitals and medical centers, it should also be considered. Another bias in decision support systems is \textit{automation bias}, which is defined as the tendency to over-trust them. This problem is often reported in medical decision support systems, where clinicians rely on the software too much and overlook contradictory information \cite{goddard2012automation}. Such data, if recorded and used to build new data collections, will affect new models.

\begin{figure}[!htb]
\centering
  \includegraphics[width=0.5\textwidth]{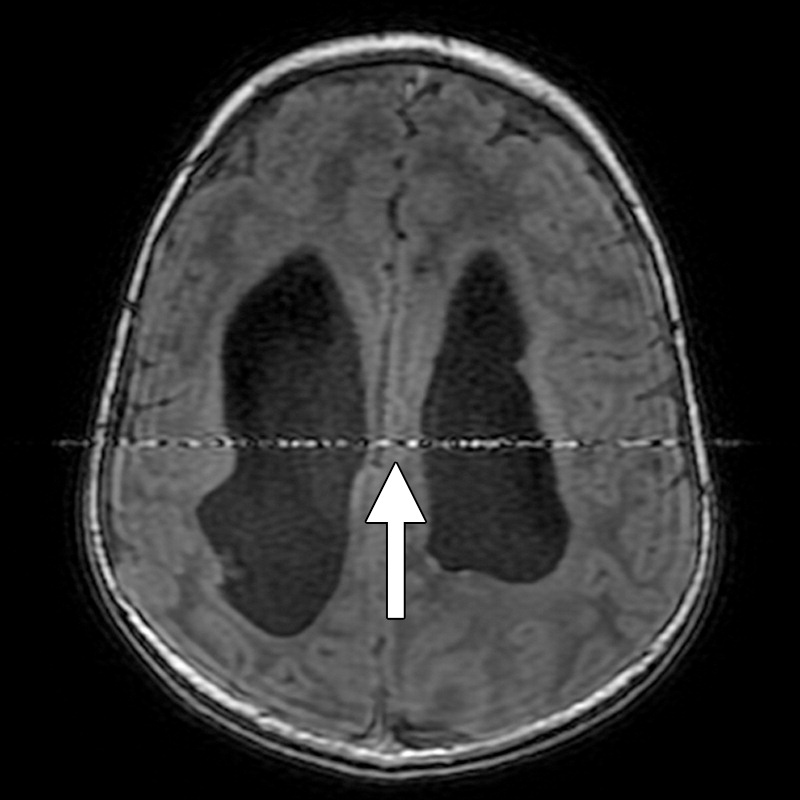}%
  \caption{Data handling bias is introduced by improper device calibration. MRI example -- zipper artifact (image source \cite{yanasak2014mr})} \label{fig.mri}
\end{figure}

Next, we have the \textit{data handling bias}. This bias describes how data is handled, which sometimes might distort the output. For instance, scanning the medical images to move them from analog format to digital might add unwanted artifacts. Nathan E. Yanasak et al. presented several domain-specific artifacts that might appear when improperly calibrating parallel magnetic resonance imaging  (MRI) \cite{yanasak2014mr}. As illustrated in Figure~\ref{fig.mri}, they include chemical shifts, zippers, ghosting, and others. The ML algorithm might wrongly consider such artifacts as an essential feature. 

Another interesting case highlighting the significance of data handling bias has been recently reported by Bobowicz et al. \cite{bobowicz2023attention}. The authors presented a case study focusing on breast cancer diagnosis through the analysis of mammographic images. Initially, they achieved unexpectedly high results, prompting them to further examine the data.

\begin{figure}[!htb]
\centering
  \includegraphics[width=0.7\textwidth]{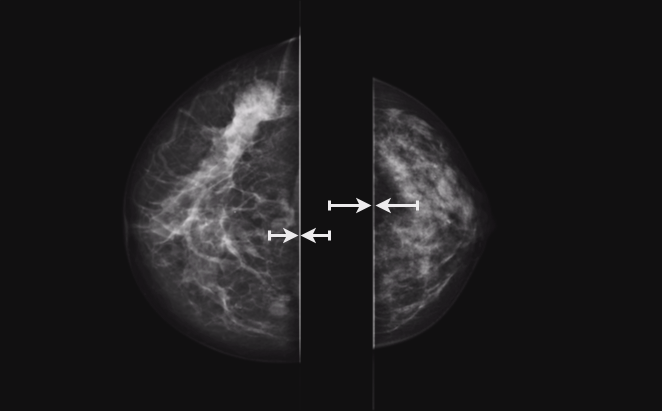}%
  \caption{Example of collimator misalignment which was correlated with the breast cancer prognosis, resulting in data handling bias.} \label{fig.collimator}
\end{figure}

During the data examination, they discovered an artifact related to collimator misalignment, characterized by the presence of high-value (white) pixels near the image edges, as illustrated in Fig. \ref{fig.collimator}. While radiologists often overlook such artifacts, they can have a significant impact on prediction outcomes. In this particular case, the presence of the collimator misalignment artifact introduced a biasing factor, resulting in an improvement in accuracy of approximately 7 percentage points.

Hence, the seemingly 'high' accuracy results were attributed to the presence of the collimator misalignment artifact, which exclusively occurred in images from breast cancer patients. Considering the characteristics of real-world medical data, where non-cancer cases typically outnumber cancer cases, the analyzed dataset covered a wider range of acquisition years for the rarer class, as confirmed by metadata inspection. The variations in acquisition times between these two classes could potentially influence the data distribution and characteristics, leading to an imbalanced representation and possible bias towards certain time periods associated with collimator calibration procedures.


Ensuring high-quality data is essential for reliable and accurate machine learning results. By identifying and addressing labeling errors and biases in the dataset, we can improve the overall performance of machine learning models.

\subsection{Data Analysis} \label{section.data_analysis}

Even when the dataset acquisition process is well-thought and well-organized, data is collected carefully, following the guidelines, the bias might still appear in Stage 3: Data analysis. Analysis bias is defined as the result of errors in data analysis.  

One of the most mentioned problems in the data analysis stage is a \textit{Confounding bias}. Confounding phenomena has been studied since the early '70s by epidemiologists, statisticians, doctors, and mathematicians \cite{mccroskey1966ethos,axelson1978aspects,greenland1980control}. Epidemiologists define a confounder as a pre-exposure variable associated with exposure and the outcome conditional on the exposure, possibly dependent on other covariates \cite{miettinen1974confounding}. 
In statistics, a \textit{confounder} (also known as a confounding variable, confounding factor, extraneous determinant, or lurking variable) is a variable that influences both the dependent variable (i.e., disease) and independent variable (the studied factor), causing a spurious association \cite{pearl2009causal,vanderweele2013definition}. 

To better understand a confounding factor, consider studying the relationship between drinking coffee daily and having heart problems \cite{tulchinsky2014measuring}. It might look like coffee causes heart problems because coffee drinkers statistically have more cardiovascular diseases. However,  coffee drinkers smoke more cigarettes than non-coffee drinkers. We might notice that smoking is a\textit{ confounding variable} in the study of the association between coffee drinking and heart disease. A higher probability of heart disease might be due to smoking rather than coffee drinking. More recent studies have shown coffee drinking to have substantial benefits in heart health and the prevention of dementia \cite{bidel2013emerging}.

\begin{figure}[!htb]
\centering
  \includegraphics[width=0.5\textwidth]{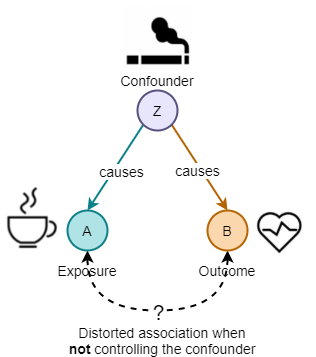}%
  \caption{Confounding bias. Smoking is an example of distorted association when studying the relationship between drinking coffee daily and having  heart problems when not controlling confounding factors.} \label{fig.confounder}
\end{figure}

According to McNamee et al., \cite{mcnamee2003confounding} \textit{confounding bias}, a systematic error can occur in epidemiological studies in measuring the association between exposure and the health outcome caused by mixing the exposure of primary interest with extraneous risk factors.
It is said that unlike selection or information bias \cite{althubaiti2016information}, confounding is one type of bias that can be adjusted after data gathering using statistical models \cite{pourhoseingholi2012control}. However, to decide whether a variable is working independently, a biological or social mechanism must cause exposure to the disease or health outcome \cite{alexander2015confounding}.
A confounding bias is a widely recognized problem in social sciences and causal modeling. In ML, it receives much less attention \cite{landeiro2017controlling}. However, the problem still exists and might remain unnoticed due to the high dimensionality of the current issues solved by deep learning. 

\begin{figure}[!htb]
\centering
  \includegraphics[width=0.6\textwidth]{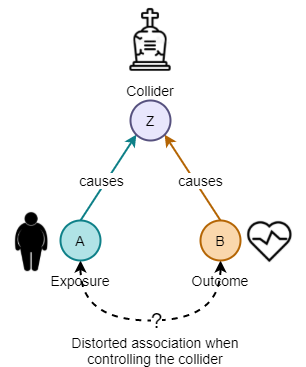}%
  \caption{The collider bias. Example of distorted association when studying the relationship between obesity and having  heart problems when controlling collider factor -- mortality rate.} \label{fig.collider}
\end{figure}

Next to the confounding bias stands \textit{collider bias}, also known as \textit{collider-stratification bias} \cite{10.2307/3703850} or \textit{reversal paradox} \cite{10.1093/aje/kwi002}). Collider bias is a causally influenced association between two or more exposures when a shared outcome (collider) is included in the model as a covariate \cite{day2016robust}. The main difference between confounding and collider bias is that confounders should be controlled when estimating causal associations, whereas colliders are not, as presented in Figure~\ref{fig.collider}). An interesting example of the collider is an obesity paradox \cite{viallon2016can}. An obesity paradox says that people with cardiovascular diseases and obesity have lower mortality rates than those without obesity. In a sample with only people with cardiovascular diseases, such observation creates a distorted association of the preventive effect of obesity on mortality. It is well known that obesity increases mortality rates \cite{viallon2016can}. In that case, the mortality rate is a collider - it is affected not only by obesity but also by other unmeasured factors.

Other frequently mentioned types of bias are \textit{analysis strategy} and \textit{post hoc analysis} biases.

\begin{figure}
\centering
  \includegraphics[width=0.9\textwidth]{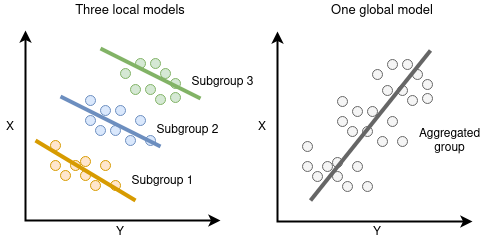}%
  \caption{The Simpson's Paradox - when the relationship between two variables differs within subgroups and its aggregation \cite{mehrabi2021survey}} \label{fig.simpsons}
\end{figure}

The \textit{Reversal paradox }(sometimes called amalgamation paradox) happens when the association between two (or more) variables can be reversed when another variable is statistically controlled for \cite{messick1981reversal}. The most known subtype of the reversal paradox is \textit{Simpson's Paradox} (Yule-Simpson effect). Simpson's paradox can be observed when the relationship between two variables differs within subgroups, and their aggregation \cite{tu2008simpson}. The Simpson's paradox is presented in Figure \ref{fig.simpsons}: the relationship between two variables  -- $X$ and $Y$ illustrated on axes --  is different for subgroups (higher $X$ means lower $Y$) and its aggregation (higher $X$ means higher $Y$).

\subsection{Model selection and training}

One of the most important steps in a research project is selecting a proper model for the corresponding problem. Even when the data is free from bias, the final predictions still might be biased. When the model is the source of bias, it is called an \textit{Algorithmic Bias} \cite{baeza2018bias}. Some sources also define an algorithmic bias as amplifying and adversely impacting existing inequities in an algorithm, e.g., socioeconomic status, race, ethnic background, religion, gender, disability, or sexual orientation \cite{panch2019artificial}. The general design based on feedback loops is also criticized widely, as researchers say that the self-reinforcing feedback might amplify the inequities \cite{nisan2001algorithmic,lum2016statistical, edelman2017racial}. The problem of bias amplification is often mentioned, e.g., in recommending engines \cite{lloyd2018bias}, word embeddings \cite{bolukbasi2016man}, or other models considered discriminate \cite{mayson2018bias}. Let us assume we have a dataset for cat vs. dog classification. In the dataset, if an animal is sitting on the grass, it is a dog in 70\% of cases. However, after training the model, the predictions showed that 85\% of animals on the grass were classified as dogs, as the grass became an essential feature for the classifier, amplifying already existing bias. The problem was highlighted by Zhao et al. in the paper \textit{Men also like shopping: Reducing gender bias amplification using corpus-level constraints} \cite{zhao2017men}. A similar problem shows a significant \textit{gender bias} in commonly used benchmark datasets. The women were more often found in the kitchen, and as a result, introduced a strong gender bias in the algorithm \cite{zhao2017men}. Stock et al. \cite{stock2018convnets} found a similar problem  connected to the uneven distribution of players' skin color in different sports in in ImageNet. Model trained to recognizing basketball, ping pong, and volleyball players focused more on the player's skin color than characteristic ball, t-shirts, or background \cite{stock2018convnets}. Also, many NLP corpora have been proven to be influenced by gender bias \cite{kaneko2022gender}. Those and other biases might be amplified by algorithmic bias.

Some even call those biased models self-fulfilling prophecies \cite{cowgill2019economics}. Currently, the algorithmic bias is often discussed due to the increasing popularity of \textit{algorithmic fairness}, i.e., the concerns that algorithms may discriminate against certain groups \cite{edelman2017racial}. It is well known that algorithms (models) can inherit questionable values from data and acquire or amplify biases during the training \cite{wong2019democratizing}. However, some researchers believe that selecting an adequate model (or training procedure) will eliminate biased predictions \cite{cowgill2019economics}. The subject of "unfair algorithms" caught public attention: numerous people shared their stories on social media (e.g., Twitter) on how they were victims of algorithmic fairness. A real-life example of a discussion on the problematic algorithmic (un)fairness would be \cite{corbett2017algorithmic} a COMPAS -- Correctional Offender Management Profiling for Alternative Sanctions. COMPAS is a decision support tool used in the US to predict recidivism risk, i.e., a criminal defendant will re-offend. The reported problem with the device is that it gave significantly higher false-positive rates against black people \cite{juliaangwin_2016}. Such calculated risk strongly affected the judge's decision. It seemed that the algorithm more often classified black people as people at a high risk of committing a crime again - and made mistakes more often \cite{juliaangwin_2016}.

\begin{table}[!htb] \label{tab.propublica}
\caption{ProPublica's table (2016) reporting model errors at the study cut point (Low vs.  Not Low) for the General Recidivism Risk Scale \cite{juliaangwin_2016}}
\centering
\begin{tabular}{llll}
\hline
\textit{COMPAS Risk Prediction} & \textit{Reoffend} & \textit{White} & \textit{Black} \\ \hline
High Risk                      & No                & 23.5\%         & 44.9\%         \\
Low Risk                       & Yes               & 47.7\%         & 28.0\%         \\ \hline
\end{tabular}

\end{table}

That controversial article claimed that almost half of black people were mistakenly classified as high-risk re-offenders. In contrast, nearly half of white people were incorrectly classified as low risk, as presented in Table~\ref{tab.compas}. Later, another publication explained why those results were wrong and presented contradictory statistics showing properly calculated statistics \cite{dieterich2016compas}. This is a great example of \textit{Simpson's Paradox}. 

\begin{table}[!htb]
\caption{Northpointe Inc. Research Department table (2016) reporting model errors at the study cut point (Low vs.  Not Low) for the General Recidivism Risk Scale \cite{dieterich2016compas}}\label{tab.compas}
\centering
\begin{tabular}{llll}
\hline
\textit{COMPAS Risk Prediction} & \textit{Reoffend} & \textit{White} & \textit{Black} \\ \hline
High Risk                      & No                & 41.0\%           & 37.0\%           \\
Low Risk                       & Yes               & 29.0\%           & 35.0\%           \\ \hline
\end{tabular}
\end{table}

Similar news and raising public awareness leads to higher demand for eliminating algorithmic bias and "fairer algorithms design." \cite{wong2019democratizing}
Hence, an \textit{algorithmic bias} might be defined as a bias that is amplified or introduced by the model.

A model can also inherit algorithmic bias. In ML, it is common to employ more than one model to perform a given task. A standard study in computer vision, detection is still often performed as a two-stage process. First, the object of interest is localized on the image, and then it is classified by the further model \cite{majchrowska2022deep}. Similarly, many action recognition models work. First, a tool for pose estimation is used, and the coordinates are passed to another model \cite{li2021ta2n}.
Moreover, trained models are sometimes used to quickly label new data, or to fine-tune models to the domain task instead of training from scratch. Sometimes pretrained models are used as feature extractors, and another algorithm takes care of the target job. If the used model is biased, the next model in the sequence can inherit these tendencies. This bias is called \textit{inherited bias}. The term was introduced by Hellstrom et al.\cite{hellstrom2020bias}. Sun \cite{sun2019mitigating} (as noted by \cite{hellstrom2020bias}) identifies several NLP tasks that may cause an inherited bias: machine translation, caption generation, speech recognition, sentiment analysis, language modeling, and word embeddings \cite{sun2019mitigating}. In \cite{sun2019mitigating} an example is presented of how different tools for sentiment analysis predict different sentiments for the same utterances but other subject's gender. In one of very recent studies \cite{mikolajczyk2022debiasing}, it was proved that generative models are vulnerable to catching and enhancing biases from data. GANs showed that they not only recreate bias in data but also significantly enhance it. For instance, one of the artifacts that was more prominent in one of the classes was never generated in the other (even though it naturally occurred in other class). Further models fed with data generated by GANs inherited those enhanced biases, resulting in even less robust models than those trained with no data augmentation at all.

Algorithmic bias can also come from the model itself. There is a noticeable \textit{learning bias} of deep networks toward low-frequency functions \cite{rahaman2019spectral}. It is well known that models prioritize learning simple patterns that generalize across data samples. Following that, Rahaman et al. \cite{rahaman2019spectral} investigated the shape of the data manifold by presenting both higher and lower frequencies to the models. They observed that lower frequencies are generally easier to learn (and learned first), and high frequencies get easier to train with increasing complexity. They called this tendency to favor smooth functions a \textit{spectral bias}. 

Recently, the image frequency in image classification was examined. It was discovered that some convolutional neural networks classify images by texture rather than by shape \cite{hermann2020origins}. The researchers examined the effect -- they experimented with mixing images with conflicting shapes and textures, e.g., a cat picture with an elephant skin texture. The CNNs tendency to classify images by texture despite particular objects' shapes was called a \textit{texture bias}. Authors showed that a random-crop data augmentation increases texture bias, and appearance-modifying data augmentation reduces it. On the other side, a \textit{shape bias} uses features primarily based on the item's shape in contrast to the texture.
 
Similarly, generative models tend to generate specific frequencies, making it easy to differentiate from the real ones \cite{schwarz2021frequency}. This tendency is called a \textit{frequency bias} \cite{schwarz2021frequency} or \textit{spatial frequency bias} \cite{khayatkhoei2020spatial}. The paper shows that generating images leaves a trace of systematic artifacts that could be recognized as fake solely by spectra analysis.

Next, there are sources that define an \textit{omitted variable bias} (OVB) \cite{cinelli2020making}. This bias arises when one or more critical variables are deliberately or unintentionally left out of the analysis. For example, in predicting stock market prices, it may be important to consider factors such as sentiment analysis on news articles about the companies valued on the market, in addition to analyzing data from previous months and years \cite{cakra2015stock}. It can be challenging to select all the essential variables that might affect the final prediction accurately.

The omitted variable bias overlaps partially with the definition of confounding bias, as they both cover the effect of omitting important variables in the analysis. However, the omitted variable bias is agnostic to the causal relationship between the variables, in contrast to the confounding bias. It means that omitting the variable in OVB will result in an unbiased estimate for the total causal effect, but in confounding bias, the total estimate will be biased. Therefore, omitting a variable may affect your predictor either in lower efficiency/quality because of not including some factor due to causally biased results (as previously mentioned confounding bias in Figure \ref{fig.confounder}), or because of not including additional important information (OVB, e.g. not including patient metadata when analyzing RTG images, when patient characteristics are essential factors).

However, sometimes not only the design or feature selection affects the final result. At the time, the model might have been considered unbiased. Yet, a few months or years later, it could be burdened with an \textit{emergent bias} \cite{friedman1996bias}. Emergent bias arises in the context of use with real users \cite{friedman1996bias}. This bias typically emerges a while after training is finished due to changing societal knowledge, population, or even cultural values. Moreover, bias can occur when used by a population with different values than those assumed in the design. An example would be predicting the risk of obesity based on somebody's living area, which might change over time due to constantly changing eating habits, health awareness campaigns, or even changes in how obesity is defined. Emergent bias can be divided into more sub-types as described in \cite{friedman1996bias}.

Improper or poorly performed evaluation can also affect a model. This type of bias is usually referred to as \textit{evaluation bias} introduced during the model's evaluation \cite{gordon1995evaluation}. The definition includes poorly selected evaluation data (e.g., inappropriate benchmarks) or inadequate metrics that do not measure the model's performance sufficiently \cite{suresh2019framework}. An example would be choosing the model based on an average accuracy in class imbalance: a model might quickly achieve 90\% accuracy by always predicting the same class when ninety percent of the whole dataset belongs to that class. Similarly, an \textit{illusion of control bias} happens when a designer achieves a high accuracy (or other metrics) and believes in controlling it \cite{masis2021interpretable}. However, high results on a test set do not always mean that model generalizes well. 
Sometimes additional measures are needed e.g. checking the behavior with outliers, or examining it with explainable AI methods.

Finally, at the deployment stage of the model preparation, a \textit{deployment bias} can occur \cite{baker2021algorithmic}. A system is used or interpreted in inappropriate ways \cite{suresh2019framework}, e.g., when a model is used for a different purpose than the initially designed purpose. Sometimes, when model is quantized for the efficiency purposes it can achieve different results than original model.

\subsection{Results interpretation }

Even if the data analysis is conducted correctly, it is still possible to misinterpret the results due to an \textit{interpretation bias}. One such bias is the \textit{correlation bias} or \textit{cause-effect bias}. This occurs when a correlation between two variables is wrongly assumed to be a causal relationship. Correlation indicates that there is a relationship or pattern between the values of two variables \cite{altman2015association}. On the other hand, causation implies that one event causes another event to occur \cite{altman2015association}. Confusing these two can lead to erroneous assumptions that can bias the outcome of the research.

\begin{figure}[!htb]
\centering
  \includegraphics[width=0.9\textwidth]{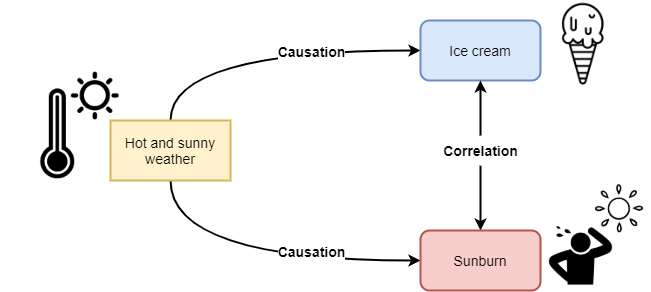}%
  \caption{Correlation bias - when the correlation between two or more variables is incorrectly mistaken with causation.} \label{fig.correlation}
\end{figure}

An example of the cause-effect bias is the "hot day" example. On hot days, people tend to buy more ice-cream and also spend more time in the sun, which increases their likelihood of getting sunburned. Consequently, there appears to be a correlation between the number of ice-creams sold and the number of people with sunburns (see Figure \ref{fig.correlation}). However, it would be erroneous to conclude that ice-cream consumption causes sunburns.

Another type of interpretation bias is the \textit{conceptual bias}, also referred to as an \textit{assumption bias}. This bias arises from flawed logic, incorrect premises, or mistaken beliefs of the researcher \cite{k2014bias}. In psychology, this is similar to the \textit{belief bias}, which is the tendency to assess an argument's validity based on personal beliefs rather than the evidence presented \cite{markovits1989belief}. 

Research on adults has shown that their performance is often influenced by empirical factors rather than the premises or the conclusion. In an early study on belief bias, Markovits et al. \cite{markovits1989belief} presented over 150 subjects with two premises that were intended to be considered as ground truth and a conclusion to label as either \textit{True} or \textit{False}. One example presented was whether cats are animals or not, as shown in Figure~\ref{fig.cats}. Surprisingly, most people tended to answer based on their personal belief rather than the premises and selected the answer as \textit{False}, even though the premises clearly indicated that cats cannot be animals if they don't like water.

\begin{figure}[!htb]
\small
\centering
    \begin{verbatim}
    
    Task: For each problem, decide if the given conclusion
    follows logically from the premises. Circle YES if,
    and only if, you judge that the conclusion can be derived
    unequivocally from the given premises, otherwise circle NO.

    Example syllogisms:
    (1) Premise 1: All animals love water.
        Premise 2: Cats do not like water.
        Conclusion: Cats are not animals. [YES/NO]
        
    (2) Premise 1: All flowers have petals.
        Premise 2: Roses have petals.
        Conclusion: Roses are flowers.\end{verbatim}
    \caption{Belief bias examination. Questionnaire testing if subjects value more the premises or personal beliefs -- example from literature \cite{markovits1989belief}}
 \label{fig.cats}
\end{figure}

\subsection{Publication}

After completing all the previous stages, the researcher may attempt to publish the results. However, bias can still be present even at this final stage. In machine learning, a new type of bias known as \textit{resubmission bias} has emerged \cite{stelmakh2021prior}. This bias can create a horn effect on a manuscript previously rejected at a different venue. The impact of resubmission bias on the overall score received by submissions appears to be small, but in highly competitive top machine learning conferences, even small changes in review scores can significantly impact the outcome. For example, the data from the ICML 2012 conference shows that papers with a mean reviewer score of 2.67/4.0 were six times more likely to be accepted than papers with a mean score of 2.33 \cite{stelmakh2021prior}.

Another common phenomenon is \textit{funding bias}, which can occur when a party reporting results does so to satisfy the research study's funding agency or financial supporter \cite{mehrabi2021survey}. 

Another problem is a \textit{presentation bias} which can results in valuable papers being omitted, or their impact being reduced due to how the research topic or information is presented \cite{mehrabi2021survey}. 

\textit{Ranking bias} suggests that top-tier journals and conferences receive much more attention than local ones, even if the quality and scope of the research are the same \cite{mehrabi2021survey}. Sometimes, the number of citations can bias users, as is the case with the incorrect analysis of COMPAS \cite{juliaangwin_2016} that was mentioned earlier (see algorithmic bias). Even though it is flawed, this research paper is still cited as a reliable source by some. Ranking bias might be considered as a subtype of a bias caused by a halo effect.

The popularity bias can extend to search engines and crowd-sourcing applications. This, in turn, can lead to \textit{position bias}, where users tend to click on higher ranked results \cite{lerman2014leveraging}. Although users tend to scan results in a particular order based on rank, clicking on higher ranked results does not necessarily imply that those results are relevant or of high quality.

\section{Bias detection}


In the previous sections, we introduced several possible biases that could inadvertently contaminate data or models. Spreading awareness about the consequences of poorly approached data or model preparation can help prevent the spread of bias. However, questions arise, what can we do when unwanted tendencies are already present in gathered data or used models? How can we ensure that our models are free from systematic errors? This is where bias detection becomes crucial.

Research on bias analysis mainly focuses on detecting causal connections between input features and predictions of trained models \cite{balakrishnan2021towards}. One approach to bias detection is manual inspection of data and models, which relies mainly on observational studies. Statistical methods can help understand complicated statistics and reveal hidden spurious correlations that may influence the model. However, manually annotating or inspecting vast datasets that fuel deep learning algorithms, such as ImageNet (with over 14 million images), Amazon Reviews (with over 82 million text reviews), and Common Voice (with over 1000 hours of speech), may not be feasible.

Local explainability methods, such as attribution maps visualized as heatmaps or visualizations based on prediction perturbation, can aid manual review. Schaaf et al. \cite{schaaf2021towards} compared different attribution maps and their ability to detect bias. They introduced and applied several metrics, such as the Relevance Mass Accuracy, Relevance Rank Accuracy Accuracy or Area over the perturbation curve (AOPC), which help evaluate the relevance of attribution maps. Large values of AOPC indicate that perturbation significantly decreases prediction accuracy, indicating that the attribution method efficiently detects relevant image regions \cite{schaaf2021towards}. The authors were able to find quantitative evidence that attribution maps can be used to detect some data biases. However, their analyses also showed that attribution maps sometimes provide misleading explanations.

\begin{figure*}[!htb] 
\centering
\footnotesize

\includegraphics[width=0.25\linewidth]{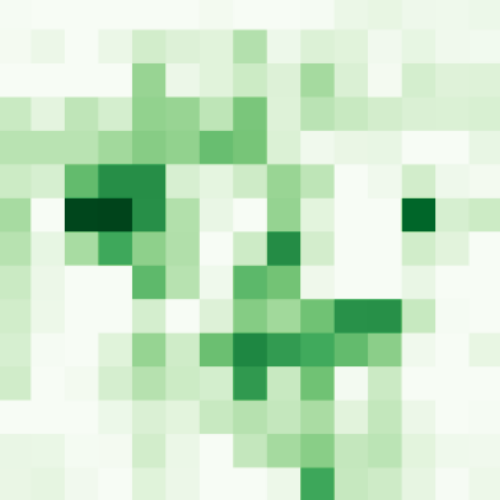}

A. Small skin lesion with smooth borders on the center of the image with strongly textured skin

    \qquad
    
        \includegraphics[width=0.25\linewidth]{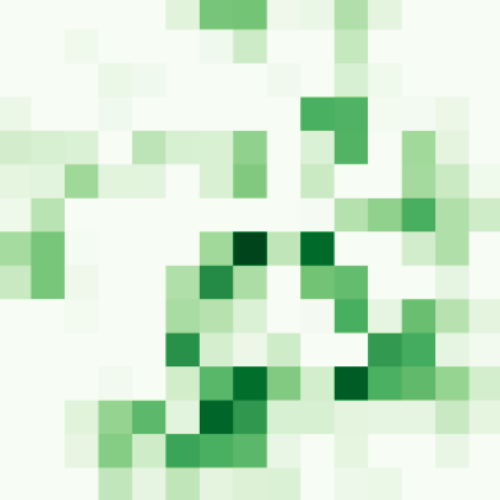}

        B. Large protruding skin lesion with well-defined borders
    \qquad
    
    \includegraphics[width=0.25\linewidth]{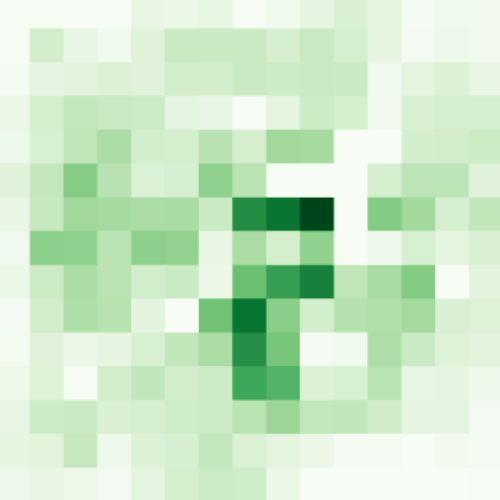}

    C. Medium round skin lesion with irregular border with streaks and atypical dots
    \qquad
  \caption{Example visualization of occlusion-based explanations. In the heatmap, a darker green color means stronger attribution. Visualized with captum \cite{kokhlikyan2020captum}.}\label{fig.gebi.attribution}
\end{figure*}

Another approach to (semi) automated bias discovery takes advantage of global explainability methods. One such method is SpRAy \cite{lapuschkin2019unmasking}. The idea behind this method is to generate attribution maps of all data instances and then cluster those maps to reveal hidden patterns in the model's reasoning. Global explanation allows users to avoid a time-consuming manual analysis of individual attribution maps but requires manual review of resulting clusters. However, there is one significant flaw with the method: it uses only the attribution maps. Analyzing the attribution maps without an input image is a challenging task. Figure~\ref{fig.gebi.attribution} presents a few examples of attribution maps of skin lesions without source images. It is difficult to discern what grabbed the attention of the prediction.

Another approach, proposed by Stock et al. \cite{stock2018convnets}, used an adversarial version of model criticism initially proposed by Kim et al. \cite{kim2016examples}, and a  feature-based explanation to uncover potential biases. Model criticism summarizes the relations between input features learned by a model using a carefully selected subset of examples (prototypes). Such a tool helps in manual data and model investigations and could be automated shortly. 

Balakrishnan et al. \cite{balakrishnan2021towards} proposed using Generative Models for building causal benchmarks. Generative models manipulate input features, e.g., gender and skin tone, to reveal potential causal links between feature variation and prediction changes. However, as the author mentions, those Generative Models are hardly controllable, and hidden confounders can still be present in benchmarks. 

Another approach proposed by Serna et al. \cite{serna2021ifbid} is an Inference-Free Bias Detection that, in contrast to other approaches, tries to detect bias by investigating the models through their weights. In the paper \cite{serna2021ifbid}, bias is detected with an additional detector model, which tries to detect bias encoded in the parameters of the trained model. The detector model takes as input weights of the trained model: the architecture depends on the weights/filters of each layer and has a dense layer that concatenates all the outputs of each weight/filter module. The performance is proved to be quite good. Still, it can be used to caution the user during the inference rather than detect new bias in training data, as it requires a training dataset to find biases.

The literature about bias detection is relatively poor. There are no guidelines or widely used algorithms to help bias discovery in models and data. The only available option is manual annotation and manual data inspecting.

\section{Bias mitigation} \label{section.bias-mitigation}


Bias mitigation methods from classical literature usually operate on simple, often linear models \cite{wang2020towards}. However, such approaches are insufficient for solving the problem in the deep learning era. It is not feasible to abandon high-efficiency models for simpler linear algorithms. However, completely ignoring possible biases in data and models is not a viable solution, as deep learning-based models are increasingly used in practice. It is well-documented that models reflect the bias in data and often amplify it \cite{zhao2017men}. Therefore, a new area of research has emerged towards mitigating biases in data and models for safer, more robust, and fairer deep models without sacrificing their size or architecture.

One approach is fairness through blindness \cite{wang2020towards}. The idea is simple: if we suspect that a variable might bias the model, we should not include it as a set of input features. For example, we might not want to include information about the candidate's gender when evaluating a job candidate's resume. However, if a model encodes information about a protected variable, it cannot be considered unbiased. In reality, some information about gender might be encoded in the resume, such as feminine hobbies or gender-specific adjectives. Removing all potential biases is often a challenging, if not impossible, task.

Therefore, other approaches have emerged. For instance, Zhao et al. \cite{zhao2017men} proposed an inference update scheme to match a target distribution to remove bias. Their method introduces corpus-level constraints so that selected features co-occur no more often than in the original training distribution. Next, Dwork et al. \cite{dwork2018decoupled} proposed a scheme for decoupling classifiers that can be added to any black-box machine learning algorithm. These can be used to learn different classifiers for different groups. Another branch is adversarial bias mitigation, where the task is to predict an output variable $Y$ given an input variable $X$, while remaining unbiased with respect to some variable $Z$ \cite{zhang2018mitigating}. The approach proposed by Zhang et al. \cite{zhang2018mitigating} uses the output layer of the predictor as an input to another model called the adversary network, which attempts to predict $Z$. The idea was improved by Le Bras et al. \cite{le2020adversarial}, who proposed the idea of Adversarial Filters of Dataset Biases - AFLite. The proposed algorithm uses linear classifiers trained on different random data subsets at each filtering phase. Then, the linear classifier's predictions are collected, and a predictability score is calculated. High predictability scores are undesirable as their feature representation can be negatively exploited. Therefore, Le Bras et al. \cite{le2020adversarial} proposed simply removing the top $n$ instances with high scores. The process is then repeated several times to reduce the bias influence.

Finally, there is attention guidance. Early works about attention guidance in computer vision focused on improving the segmentation task \cite{huang2019brain}, making classification better with attention approaches used in Natural Language Processing \cite{barata2019deep}, or even using attention maps to zoom closer to the region of interest \cite{li2019zoom}. However, attention in terms of vision transformers is different from attribution maps. Researchers argue that they can be used interchangeably. In 2019, Jain et al. \cite{jain2019attention} explained why they think that "Attention is not an explanation", whereas the authors of \cite{wiegreffe2019attention} claimed that \textit{Attention is not not an explanation}. The conflict did not die with time, as many other papers regarding this matter appeared \cite{grimsley2020attention,tutek2020staying,woody2004more}.

One of the similar emerging approaches is attention guidance \cite{tang2018attention}. The guidance provided with, for example, attention maps highlights relevant regions and suppresses unimportant ones, enabling a better classification. Self-erasing networks are also based on a similar method that prohibits attention from spreading to unexpected background regions by erasing unwanted areas (Hou et al., 2018). Different solutions have been proposed by researchers to address this problem, such as rule extraction, built-in knowledge, or built-in-graphs (Chai et al., 2020). However, there is still a long way to go to achieve full transparency of the reasoning process of DNNs and incorporate it into the training process.

\section{Conclusions} 
Currently, bias in machine learning research is often discussed in terms of fairness, while overlooking the original definition of bias as a "systematic error". We showed that this has created a gap between past and current research on bias, particularly in terms of understanding its root causes and negative effects which makes bias mitigation challenging. In this paper, we showed previous and current research works on machine learning with a goal of closing the gap between them. We presented over forty potential sources of bias in the machine learning pipeline: literature review, data collection, data analysis, model selection and training, interpretation of the results, publication. Furthermore, we briefly presented methods for detecting and mitigating bias, including fairness metrics, debiasing techniques, and explainability methods.

\section{Acknowledgements} 
The  research on bias reported  in  this  publication  was supported  by  Polish  National  Science  Centre (Grant Preludium No: \textit{UMO-2019/35/N/ST6/04052}).

\vspace*{-.5pc}

\enlargethispage*{\baselineskip}
\bibliographystyle{plainnat}
\bibliography{Bibliography}


\end{document}